*Original Article*

# A Novel Multimodal Framework for Early Detection of Alzheimer's Disease Using Deep Learning

Tatwadarshi P. Nagarhalli[1], Sanket Patil[2], Vishal Pande[3], Uday Aswalekar[4], Prafulla Patil[5]

[1,2,3,4,5]*Vidyavardhini's College of Engineering and Technology, University of Mumbai, Mumbai, India.*

[1]*Corresponding Author : tatwadarshipn@gmail.com*

***Abstract*** - *Alzheimer's Disease (AD) is a progressive neurodegenerative disorder that poses significant challenges in its early diagnosis, often leading to delayed treatment and poorer outcomes for patients. Traditional diagnostic methods, typically reliant on single data modalities, fall short of capturing the multifaceted nature of the disease. In this paper, we propose a novel multimodal framework for the early detection of AD that integrates data from three primary sources: MRI imaging, cognitive assessments, and biomarkers. This framework employs Convolutional Neural Networks (CNN) for analyzing MRI images and Long Short-Term Memory (LSTM) networks for processing cognitive and biomarker data. The system enhances diagnostic accuracy and reliability by aggregating results from these distinct modalities using advanced techniques like weighted averaging, even in incomplete data. The multimodal approach not only improves the robustness of the detection process but also enables the identification of AD at its earliest stages, offering a significant advantage over conventional methods. The integration of biomarkers and cognitive tests is particularly crucial, as these can detect Alzheimer's long before the onset of clinical symptoms, thereby facilitating earlier intervention and potentially altering the course of the disease. This research demonstrates that the proposed framework has the potential to revolutionize the early detection of AD, paving the way for more timely and effective treatments.*

## 1. Introduction

Alzheimer's Disease (AD) is a neurological illness that worsens with time and is typified by a loss of memory, cognitive function, and daily functioning. It is one of the most common disorders affecting the elderly globally, and despite a great deal of study, there is still no cure. On the other hand, early discovery of AD might potentially slow down the disease's course and lessen its effects, improving the lives of people who are impacted [1].

The cognitive impairments caused by AD, particularly in abstract thinking and problem-solving, can severely disrupt daily life. Tasks such as managing finances, balancing checkbooks, and making timely payments become increasingly difficult as the disease advances. Eventually, the ability to comprehend and utilize numerical concepts may be lost [2].

Although dementia and Alzheimer's disease are frequently used synonymously, dementia is really a symptom of Alzheimer's disease. Another widespread misperception is the false belief that Alzheimer's disease develops naturally as people age. Currently, 6.2 million Americans, 65 and older, have Alzheimer's dementia; if no significant medical advancements occur, this figure might increase to 13.8 million by 2060. According to official death certificates, 121,499 fatalities in the US in 2019 were directly related to Alzheimer's disease, making it the sixth most common cause of death overall. Interestingly, mortality from AD rose by more than 145% between 2000 and 2019 despite a drop in deaths from other important illnesses, including heart disease, HIV, and stroke. An estimated 15.3 billion hours of care, worth around $256.7 billion, were given to people with Alzheimer's or other dementias in 2020 alone by over 11 million family members and unpaid caregivers [3].

The situation is equally worrying in India. The Dementia India Report 2010 [4] from the Alzheimer's and Related Disorders Society of India (ARDSI) states that there were around 3.7 million people with dementia in 2010, and that figure is projected to almost triple to 7.6 million by 2030. The documented instances in the US and India draw attention to a far bigger problem: millions more cases can go unreported and untreated, especially in impoverished and rural areas where there is a lack of knowledge about Alzheimer's disease and its early symptoms.

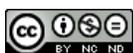

The significance of these data emphasizes how urgently increased research on AD early detection is needed. Given the irreversible nature of the condition, early intervention is critical. Several sectors, including the medical sciences, have undergone radical change due to the introduction of technologies like machine learning and deep learning, which provide significant opportunities for earlier and more accurate diagnoses.

This paper proposes a novel multimodal framework for early AD detection, integrating three key data modalities: Medical Imaging, Cognitive Tests, and Biomarker Data. By combining these diverse data sources, the framework aims to improve the accuracy of early diagnosis, providing a comprehensive approach that leverages the strengths of each modality. Medical imaging techniques like MRI and PET scans will be used to detect structural and functional changes in the brain [5, 6], cognitive tests will assess language, memory, and executive function [7], while biomarker data will offer insights into the biological processes underlying the disease [8]. This multimodal approach holds the potential to identify AD at its earliest stages, paving the way for timely interventions that could alter the course of the disease.

## 2. Literature Review

The only way to slow the growth of AD, a neurological disease that progresses and is now incurable, is to catch it early. The early diagnosis of AD has greatly benefited from recent developments in computer technology, including machine learning, deep learning, and natural language processing. Examining the efficacy of these technologies in the early identification of AD, this literature review is divided into three main data modalities: biomarkers, cognitive tests, and medical imaging.

### 2.1. Medical Imaging

The early diagnosis of AD now relies heavily on medical imaging. Numerous imaging techniques, such as Positron Emission Tomography (PET) and Magnetic Resonance Imaging (MRI), offer vital information on the anatomical and functional alterations in the brain linked to AD.

Deep Convolutional Neural Networks (CNNs) were used by Saman et al. [9] to analyze MRI data for AD classification. Their method entailed using massive MRI image datasets to train a CNN model in order to find patterns linked to the pathophysiology of Alzheimer's disease. According to the study, CNNs outperform conventional machine learning models that call on human feature extraction, with accuracy rates approaching 90%. This work highlights the potential of deep learning to improve and automate the AD diagnosis process.

The integration of multimodal imaging data, namely the combination of MRI and PET scans with machine learning methods like Support Vector Machines (SVMs), was investigated by Jack et al. [10]. The study demonstrated that the detection of tau proteins and amyloid plaques, which are indicators of AD, may be greatly enhanced by using this combination strategy. The study's significant improvements in sensitivity and specificity when employing SVMs indicate that multimodal data fusion is a viable option for early diagnosis.

Weiner et al. [11] performed a longitudinal examination of MRI scans using information from the AD Neuroimaging Initiative (ADNI). Using machine learning techniques, they could monitor changes in brain volume over time and establish a correlation between these changes and the onset of AD. According to the study, some atrophy patterns in parts of the brain, such as the hippocampus, may act as early AD markers. This study underlined the need for ongoing observation in clinical settings and proved the usefulness of longitudinal data in forecasting the course of disease.

Helaly et al. [12] introduced a deep learning framework to analyze functional MRI (fMRI) data, focusing on the brain's connectivity patterns. Their study revealed that deep learning models could detect subtle functional abnormalities in brain networks associated with Alzheimer's, even before significant cognitive decline is apparent. The findings suggest that fMRI, combined with advanced deep learning techniques, could play a crucial role in early AD detection by identifying functional disruptions that precede structural changes.

Krüger et al. [13] combined Voxel-Based Morphometry (VBM) with deep CNN to detect brain atrophy patterns specific to AD. The study highlighted the effectiveness of VBM in capturing gray matter density changes, which are crucial in distinguishing between Alzheimer's patients and healthy individuals. The CNN-based classification achieved high accuracy, particularly in identifying early-stage AD, indicating the potential of VBM as a diagnostic tool when paired with machine learning.

Therefore, there is much promise for early AD identification when deep learning and machine learning are applied to medical imaging modalities like PET and MRI. However, the availability of sizable, high-quality datasets and the standardization of imaging processes are prerequisites for these technologies' efficacy. Deep learning models are still difficult to comprehend and difficult to generalize to other populations. The most reliable solutions could come from multimodal methods incorporating clinical data and many imaging modalities, but they still need more clinical validation.

### 2.2. Cognitive Tests
Cognitive evaluations are conventional methods for AD diagnosis. These assessments look at various cognitive abilities, including language, memory, and executive function all of which are frequently compromised in the early stages of AD. Recent developments in natural language processing and machine learning have improved the interpretation of cognitive test data, resulting in earlier and more accurate diagnoses.

Irfan et al. [14] employed ensemble techniques like random forests, a machine learning technique, to analyze cognitive test scores to detect cognitive decline associated with AD early. Their study demonstrated that machine learning could identify subtle patterns in cognitive performance indicative of early AD, even before clinical symptoms become apparent. The voting mechanisms offered in the ensemble techniques showed superior predictive power compared to traditional statistical methods, suggesting that ML can enhance the sensitivity and specificity of cognitive assessments.

Kulvinder Panesar and María Beatriz Pérez Cabello de Alba [15] utilized Natural Language Processing (NLP) to analyze speech and language patterns from cognitive assessments. Their study focused on the language used by patients during standard cognitive tests, identifying linguistic markers of cognitive decline. NLP models detected changes in speech patterns that correlate with early Alzheimer's, such as increased use of filler words and reduced sentence complexity. This approach highlights the potential of NLP to uncover subtle language impairments that might not be evident through traditional scoring methods.

The Free and Cued Selective Reminding Test (FCSRT), a cognitive test created especially to identify early AD, was created by Grober et al. [16]. After using machine learning algorithms to the test data, it was shown that distinct memory retrieval patterns may differentiate between those with early AD and those in good health. The results of this study highlight the value of customized cognitive testing and Machine Learning (ML) for analyzing intricate memory functions, which are frequently impaired in Alzheimer's patients.

Petersen et al. [17] conducted a longitudinal cognitive data study to predict the progression from Mild Cognitive Impairment (MCI) to AD. By applying machine learning models, the research demonstrated that certain cognitive test scores could predict the likelihood of developing AD years before clinical diagnosis. This work underscores the potential of longitudinal cognitive testing and machine learning for early risk assessment and intervention.

In order to increase diagnosis accuracy, Albert et al. [7] combined machine learning algorithms with a variety of cognitive assessments, including the Mini-Mental State Examination (MMSE) and the Montreal Cognitive Assessment (MoCA). According to the study, machine learning models outperformed conventional techniques in diagnosing patients by better handling the complexity of cognitive data. This integration demonstrates how useful it is to improve early detection efforts by integrating cognitive testing with sophisticated analytics.

So, when paired with machine learning and NLP, cognitive tests significantly improve early AD detection. Machine learning enhances the ability to identify subtle cognitive changes that may precede clinical symptoms, while NLP provides new tools for analyzing language-related impairments. However, the effectiveness of these methodologies is highly dependent on the consistency of test administration and the quality of the data collected. Furthermore, developing predictive models requires large longitudinal datasets, which may not always be available, particularly in underrepresented populations.

Deep Convolutional Neural Networks (CNNs) were used by Saman et al. [9] to analyze MRI data for AD classification. Their method entailed using massive MRI image datasets to train a CNN model in order to find patterns linked to the pathophysiology of Alzheimer's disease. According to the study, CNNs outperform conventional machine learning models that call on human feature extraction, with accuracy rates approaching 90%. This work highlights the potential of deep learning to improve and automate the AD diagnosis process.

### 2.3. Biomarkers

Biomarkers, including Cerebrospinal Fluid (CSF) and blood-based markers, provide crucial biological indicators of AD. Applying machine learning and deep learning to biomarker data has shown promise in improving the early detection and diagnosis of AD.

Using machine learning algorithms to analyse CSF biomarkers, including tau and amyloid-beta (A$\beta$), was reviewed by Blennow et al. [18]. Their research revealed that Machine Learning (ML) algorithms could reliably categorize people according to their biomarker profiles, offering a non-invasive technique for early diagnosis. Machine learning's promise for therapeutic applications is demonstrated by its ability to interpret intricate biomarker data and recognize patterns suggestive of Alzheimer's disease.

Hampel et al. [19] explored the potential of blood-based biomarkers for Alzheimer's detection, applying deep learning techniques to analyze plasma levels of A$\beta$ and tau proteins. The study demonstrated that deep learning models could predict the presence of AD with high accuracy, offering a less invasive alternative to CSF analysis. This work suggests that blood biomarkers and advanced computational techniques could play a significant role in early diagnosis, particularly in large-scale screening efforts.

In order to improve diagnosis accuracy, Gaeta et al. [20] integrated CSF biomarkers with amyloid PET imaging using machine learning methods. Several biomarker types were integrated to offer a more thorough evaluation of Alzheimer's pathology, which allowed for earlier and more precise identification. The study emphasizes how multimodal biomarker analysis might enhance the quality of diagnoses.

Wang et al. [21] applied learning models to analyze Neurofilament Light chain (NfL) levels in both CSF and blood plasma. Elevated NfL levels were found to correlate strongly with neurodegeneration and cognitive decline, making NfL a promising biomarker for early Alzheimer's detection. The study demonstrated that learning models could effectively integrate NfL data with other clinical variables, achieving high predictive accuracy. This research underscores the value of NfL as a biomarker in detecting early neurodegenerative changes associated with Alzheimer's.

Mann et al.'s [22] investigation on proteomics—a comprehensive analysis of proteins—in discovering AD biomarkers. The study analyzed complicated protein data from blood samples using machine learning methods. The study showed the promise of proteomics in early disease identification by finding certain protein signatures linked to AD. Classification individuals based on their proteomic profiles using machine learning models, such as random forests and support vector machines, achieved significant diagnostic accuracy.

Winchester et al. [23] investigated the use of metabolomics, which studies small molecules known as metabolites, as biomarkers for AD. They applied machine learning techniques to analyze metabolomic data, identifying metabolic patterns that could differentiate between AD patients and healthy controls. The study showed that changes in specific metabolites could serve as early indicators of Alzheimer's, and machine learning models were instrumental in detecting these patterns. This approach highlights the potential of metabolomics in providing a non-invasive method for early diagnosis.

Heba M. AL-Bermany and Sura Z. AL-Rashid [24] focused on gene expression data to identify genetic biomarkers for AD. The study applied deep learning models to large datasets of gene expression profiles, identifying key genes associated with AD risk. By analyzing gene expression patterns, the research demonstrated that deep learning could predict the likelihood of developing Alzheimer's at an early stage, offering a personalized approach to diagnosis. This work illustrates the potential of integrating genetic data with machine learning for precise and early AD detection.

AlMansoori et al. [25] suggested a multi-biomarker strategy that included information from blood and clinical feature biomarkers to enhance AD identification. By using machine learning models to assess the entire data, the team was able to achieve a diagnosis accuracy that was higher than that of any one biomarker strategy. This study emphasizes how crucial it is to employ a multimodal approach to improve the early diagnosis of AD. This approach involves analyzing many biomarker types simultaneously utilizing cutting-edge computer tools.

So, biomarkers provide critical insights into the biological processes underlying AD, and when combined with machine learning and deep learning, they offer significant potential for early detection. However, the effectiveness of these methodologies is influenced by several factors, including the variability in individual biomarker levels and the need for large, well-annotated datasets to train robust models. Integrating multiple biomarkers to form a comprehensive diagnostic profile remains challenging but promises more accurate and early diagnosis. The use of advanced computational models in analyzing biomarker data is still in the early stages, and more research is needed to validate these approaches in clinical settings.

*2.4. Analysis*

Table 1 summarizes the key methodologies, findings, and critical analyses, providing a clear overview of the effectiveness and challenges of different approaches in early AD detection.

Table 1. Analysis of medical imaging, cognitive tests and biomarker techniques

| Category | Paper | Methodology | Key Findings | Critical Analysis |
|---|---|---|---|---|
| Medical Imaging | Saman et al. [9] | Deep Learning (CNNs) | High accuracy in detecting structural changes in the brain through MRI images. | It effectively identifies early AD stages but requires large datasets and computational resources. |
| | Jack et al. [10] | Machine Learning (SVM, RF) | PET scans showed metabolic decline predictive of AD. | PET scans are effective but expensive and less accessible for routine screening. |
| | Weiner et al. [11] | Hybrid Models (ML+DL) | The combination of MRI and PET improved the classification accuracy of AD. | Combining modalities improves results but increases complexity and cost. |
| | Helaly et al. [12] | Transfer Learning | Used pre-trained models for MRI data, achieving good | Transfer learning mitigates data scarcity but may lead to overfitting if |

| | | | performance with limited labeled data. | mishandled. |
|---|---|---|---|---|
| | Krüger et al. [13] | Deep CNN | Used Deep CNN to enhance MRI-based AD detection. | CNN gives good accuracy but requires careful model selection and validation. |
| Cognitive Tests | Irfan et al. [14] | Ensemble techniques for analyzing cognitive test scores | Detected subtle cognitive patterns indicative of early AD; Ensemble techniques outperformed traditional methods. | Ensemble techniques enhance the sensitivity and specificity of early AD detection. However, the model's complexity and need for extensive data might limit its accessibility and generalizability. |
| | Kulvinder Panesar and María Beatriz Pérez Cabello de Alba [15] | Natural Language Processing (NLP) for speech and language analysis | Identified linguistic markers of early AD, such as filler words and reduced sentence complexity. | NLP provides unique insights into early cognitive decline through linguistic analysis, but its effectiveness depends on the quality and variability of language data available, which could limit its broader application. |
| | Grober et al. [16] | Free and Cued Selective Reminding Test (FCSRT) + Machine Learning | Machine learning identified memory retrieval patterns distinguishing early AD from healthy individuals. | Tailored cognitive tests like FCSRT and machine learning offer targeted AD detection. However, focusing on specific cognitive domains might miss other early signs of AD, requiring complementary assessments. |
| | Petersen et al. [17] | Longitudinal Cognitive Data + Machine Learning | Cognitive test scores predicted progression from MCI to AD years before clinical diagnosis. | Longitudinal data analysis is valuable for early prediction, but the reliance on long-term data collection may delay actionable insights, necessitating faster methods for early intervention. |
| | Albert et al. [7] | MMSE, MoCA + Machine Learning | ML models provide more reliable diagnoses by handling complex cognitive data better than traditional methods. | Integrating multiple cognitive tests with machine learning enhances diagnostic accuracy. However, the potential for overfitting and the challenge of selecting the right combination of tests may limit its practical application. |
| Biomarkers | Wang et al. [21] | Machine Learning | NfL levels in CSF and plasma correlated with early neurodegeneration. | It is a promising biomarker, but variability in NfL levels may affect accuracy. |
| | Mann et al. [22] | Machine Learning | Proteomics data is used to identify protein signatures associated with AD. | Proteomics is robust but requires advanced tools and large datasets. |
| | Winchester et al. [23] | Machine Learning (SVM) | Metabolomic patterns identified as early indicators of AD. | Metabolomics offers non-invasive diagnosis but is still in the early research stages. |
| | Heba M. AL-Bermany and Sura Z. AL-Rashid [24] | Deep Learning | Gene expression data was analyzed to predict AD risk, showing early detection potential. | Gene expression data provides personalized insights but is expensive and data intensive. |
| | AlMansoori et al. [25] | Multi-Biomarker Integration | Combined blood and clinical features biomarkers, achieving high diagnostic accuracy. | Integration of multiple biomarkers is promising but complex and resource intensive. |

The literature review and the analysis table underscore the significant advances in applying machine learning, deep learning, and natural language processing across key modalities: Medical Imaging, Cognitive Tests, and Biomarkers. Each modality has demonstrated potential in identifying early signs of AD, yet significant drawbacks persist when these methods are employed in isolation. Medical Imaging, while effective in capturing structural and functional brain changes, suffers from a dependence on large, high-quality datasets and challenges in model interpretability. Although improved by machine learning and NLP, cognitive tests often fail to capture the full spectrum of AD pathology. The need for consistent administration and extensive longitudinal data limits them. Biomarkers provide critical biological insights but face variability across individuals and complexity in integration, making it difficult to establish universal diagnostic criteria.

These limitations highlight the inadequacies of relying solely on any modality for early AD detection. AD's inherent complexity and heterogeneity require a more comprehensive approach integrating multiple data sources. A multimodal approach, combining Medical Imaging, Cognitive Tests, and Biomarkers, offers the most promising path forward. By leveraging the strengths of each modality, such an approach can provide a more accurate, robust, and early diagnosis, ultimately leading to more effective interventions and better management of AD.

## 3. Proposed Multimodal Framework

A flowchart depicting a multimodal system intended for the early identification of AD is shown in Figure 1. This system integrates three different forms of data: biomarkers, imaging data, and cognitive data.

Processing is done on these data categories using specific Deep Learning (DL) or Machine Learning (ML) techniques. The final prediction about the existence or stage of AD is then created by combining the separate outputs from these models.

This framework has 4 components: working with Imaging data, working with Cognitive data, working with Biomarker data and performing aggregation.

### 3.1. Imaging Data
#### 3.1.1. Dataset
2 popular datasets can be used: OASIS-1 Dataset [26] and ADNI Dataset [27].

The Open Access Series of Imaging Studies (OASIS) includes the OASIS-1 Dataset. It offers a compilation of information about brain imaging, especially MRI scans, which are essential for studying Alzheimer's disease. Cross-sectional MRI data for 416 participants, ranging in age from 18 to 96, with mild to moderate AD and cognitively healthy persons are included in the OASIS-1 dataset.

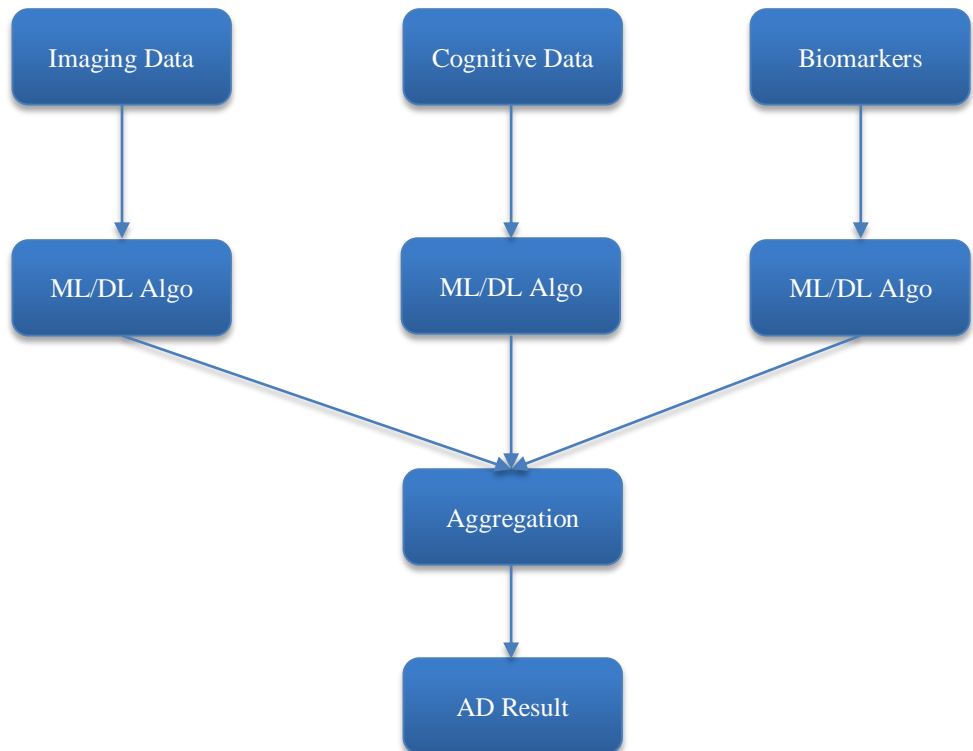

**Fig. 1 Multimodal AD detection framework**

Another popular dataset for MRI imaging data utilized in Alzheimer's research is the AD Neuroimaging Initiative (ADNI). ADNI offers longitudinal MRI scans crucial for tracking the disease's evolution. Longitudinal MRI pictures are provided. In order to train models that can anticipate Alzheimer's based on alterations in brain structure throughout time, it is utilized to investigate the course of the illness.

*3.1.2. ML/DL Algorithms*
Any ML or DL algorithms can be used. But, as the dataset contains MRI data (images), CNNs are the most effective for feature extraction and pattern recognition in imaging like Convolutional Neural Networks (CNNs). Also, since MRI data is volumetric, 3D CNNs can be employed to capture spatial features across all three dimensions.

The imaging data helps identify structural brain changes that correlate with the progression of Alzheimer's, such as atrophy in the hippocampus.

*3.2. Cognitive Data*
*3.2.1. Dataset*
Two important datasets that can be used for cognitive analysis are the OASIS-3 Dataset [28] and the NACC Dataset [29].

The OASIS-3 dataset adds biomarker, clinical, longitudinal imaging, and cognitive data, building upon previous OASIS releases. This dataset contains neuropsychological tests and cognitive evaluations like the Mini-Mental State Examination (MMSE).

Test results are among the organized cognitive data it includes. OASIS-3 monitors cognitive deterioration over time and trains algorithms that utilize cognitive performance to forecast when AD may manifest.

A dataset comprising a diverse variety of cognitive test results from individuals across several AD Research Centers (ADRCs) is made available by the National Alzheimer's Coordinating Center (NACC).

Data from organized cognitive tests is also included. This dataset is frequently used to investigate the connection between cognitive decline and illness and to create models that predict Alzheimer's based on cognitive evaluations.

*3.2.2. ML/DL Algorithms*
You may use any ML or DL method that works with time series data. However, models that perform well with time-series data, such as Recurrent Neural Networks (RNNs) or Long Short-Term Memory (LSTM) networks, can be used to interpret the cognitive data. These models may examine trends in cognitive deterioration over time, providing information about how Alzheimer's disease develops.

The cognitive data analysis helps track the decline in cognitive functions, a hallmark of AD.

*3.3. Biomarker Data*
*3.3.1. Dataset*
Two popular Biomarker analysis datasets are ROSMAP [30] and ADNI Biomarkers [31].

The Religious Orders Study and Memory and Aging Project (ROSMAP) dataset includes a wide range of data, including biospecimens, clinical and cognitive data, and detailed postmortem brain analysis. The biomarker data includes amyloid-beta, tau proteins, and other indicators relevant to Alzheimer's. The dataset contains structured biomarker data. ROSMAP is used to study AD's biochemical aspects and develop models that can predict the disease based on biomarker levels. It is particularly useful for understanding the molecular mechanisms underlying Alzheimer's.

The ADNI dataset also provides extensive biomarker data from blood and Cerebrospinal Fluid (CSF), including levels of amyloid-beta and tau proteins. This dataset is key for researchers to identify early biochemical changes associated with Alzheimer's. It contains structured biomarker data. Similar to the ROSMAP dataset, ADNI biomarker data is used to understand the biochemical pathways of Alzheimer's and to develop predictive models based on early molecular changes.

*3.3.2. ML/DL Algorithms*
Structured data from biomarkers can be analyzed using Support Vector Machines (SVMs) or Ensemble Methods like Random Forests or DL algorithms like LSTM. These models are adept at handling high-dimensional structured data to classify the presence or risk of Alzheimer's.

Biomarkers offer a biochemical snapshot of Alzheimer's progression, providing crucial information that may not be evident in imaging or cognitive data.

*3.4. Aggregation of Results*
The aggregation step combines the predictions from each of the three modalities (Imaging, Cognitive, and Biomarkers) to produce a final result regarding the presence or stage of AD.
- Weighted Average: Assigns different weights to each modality based on its importance or accuracy.
- Majority Voting: If each modality provides a binary classification (e.g., AD vs. non-AD), the majority class among the modalities can be taken as the final result.
- Stacked Generalization (Stacking): A meta-learner learns how to integrate the predictions of the various models in the best possible way by combining their outputs

- Bayesian Inference: Probabilistic models can combine predictions with uncertainties to make a more informed final decision.

The aggregation step's objective is to ensure the system is resilient if one modality malfunctions or produces incomplete data. The combined data from all accessible sources creates the final forecast or AD Result.

An inventive method for detecting AD early on is the multimodal framework, which combines biomarkers, cognitive tests, and imaging data to provide a comprehensive diagnostic tool. Diverse data formats can be used to record the diverse ways that AD presents, including biochemical changes, cognitive decline, and structural abnormalities in the brain. By examining these several data sources, the framework offers a multifaceted perspective on the illness, making it more successful than single-mode analysis at spotting early indicators of Alzheimer's.

One of the key strengths of this framework lies in its ability to gather complementary insights from different modalities. Imaging data, such as MRI scans, can reveal physical changes in the brain that are characteristic of Alzheimer's. Cognitive tests track the decline in mental functions over time, offering another window into the disease's progression. Biomarkers, conversely, can detect biochemical processes associated with Alzheimer's, which might not be visible through imaging or cognitive tests alone. By combining these insights, the framework increases the chances of early detection, as subtle changes in one modality might signal the onset of the disease before significant symptoms appear in another.

The framework's design also enhances diagnostic accuracy by aggregating results from the different modalities. This aggregation reduces the likelihood of false negatives, where the disease goes undetected, and false positives, where a patient is incorrectly diagnosed with Alzheimer's. The system is equipped to handle discrepancies between modalities, such as when one suggests the presence of Alzheimer's while others do not. In such cases, the system can flag the discrepancy and apply appropriate measures to address it, ensuring a more accurate diagnosis.

Another crucial aspect of this framework is its robustness and flexibility. The system is designed to remain functional even if one type of data is missing or of poor quality. For instance, if imaging data is unavailable, the framework can still rely on cognitive tests and biomarkers to make a prediction, although it will indicate reduced confidence in the result due to the missing data. This feature ensures the system remains operational and provides valuable diagnostic information, even in less-than-ideal conditions.

Importantly, including biomarkers and cognitive tests in the framework enables the possibility of detecting AD at a much earlier stage than traditional methods. Biomarkers, in particular, can indicate the presence of Alzheimer's years before clinical symptoms become evident. Similarly, cognitive tests can reveal early, subtle declines in mental function that might go unnoticed in everyday life. By incorporating these early indicators, the framework offers the potential to diagnose Alzheimer's much earlier, allowing for earlier intervention and treatment, which can significantly slow the disease's progression.

So, it can be said that the proposed multimodal framework represents a sophisticated and comprehensive approach to AD detection. By integrating diverse data sources, it capitalizes on the strengths of each modality to form a robust, accurate, and early diagnosis tool.

The ability to aggregate results from different modalities enhances diagnostic accuracy and ensures the system remains functional and reliable even in the face of incomplete data. This integrated approach is critical in the battle against Alzheimer's, where early detection can dramatically improve the effectiveness of treatments and interventions.

## 4. Experimentations
This section showcases how the proposed framework works.

### 4.1. Dataset
Three standard datasets have been used for the three modes proposed for early AD detection.

#### 4.1.1. OASIS-1 Dataset for Medical Imaging [26]
The scientific community can access neuroimaging datasets for AD and brain aging research through the Open Access Series of Imaging Studies (OASIS) initiative. 80000 Cross-sectional MRI scans from 416 people, ages 18 to 96, make up the OASIS-1 dataset. This dataset includes people with AD diagnoses as well as those who are cognitively normal.

High-resolution Magnetic Resonance Imaging (MRI) images have been widely employed in brain structure analysis studies to find patterns linked to aging and neurodegenerative disorders such as Alzheimer's. Researchers may associate brain structures with cognitive performance and illness development using the dataset, including diagnostic data, clinical cognitive evaluations, and demographic information.

This dataset mainly comprises neuroimaging data, specifically high-resolution MRI scans of the brain, which require specialized techniques like Convolutional Neural Networks (CNNs) for processing and analysis.

*4.1.2. OASIS-3 Dataset for Cognitive Tests [28]*
An expansion of the OASIS study, the OASIS-3 dataset provides an extensive set of longitudinal neuroimaging, cognitive, and clinical data. Data from more than 1,000 subjects, aged 42 to 95, who were observed across several visits, are included in this dataset, which is around 2800 in size. Cognitively normal people, those with Moderate Cognitive Impairment (MCI), and people with an AD diagnosis are all included in the OASIS-3 trial. A range of cognitive test results, including the Mini-Mental State Examination (MMSE) and further evaluations of executive function and memory, are included in the dataset. OASIS-3 is a complete resource for researching the course of cognitive decline and the shift from normal aging to AD since it also offers MRI and PET scans and genetic and biomarker data.

This dataset includes image data (MRI and PET) and structured data (cognitive test scores, clinical assessments). The structured data is similar to datasets like the Iris dataset but has a more complex structure due to multiple time points and different cognitive tests.

*4.1.3. ROSMAP Dataset for Biomarkers [28]*
The ROSMAP dataset combines two longitudinal cohort studies: the Religious Orders Study (ROS) and the Rush Memory and Aging Project (MAP). Clinical, pathological, and genetic data from more than 3,000 participants—many of them older individuals at risk of AD—are included in this collection. The ROSMAP dataset is especially rich in biomarker data, such as blood-based biomarkers like tau proteins and amyloid-beta (Aβ) and Cerebrospinal Fluid (CSF), which are essential to comprehend the molecular basis of AD. A thorough examination of the variables causing AD is made possible by the dataset's inclusion of gene expression data, proteomics, and thorough cognitive tests.

The ROSMAP dataset primarily contains structured data, including numerical and categorical data related to biomarkers, genetics and cognitive test scores. This dataset resembles traditional structured datasets like the Iris dataset but is much richer and more complex due to the variety of biological and clinical measurements it includes.

*4.2. Architecture*
A multimodal framework for AD early detection is shown in the flowchart in Figure 2. In order to predict the existence of AD, this framework combines three different types of data: biomarker data, MRI images, and cognitive test data that have been processed using specific machine learning or deep learning techniques.

*4.2.1. MRI Images (OASIS-1 Dataset) with CNN Data Input*
MRI scans from the OASIS-1 Dataset detect anatomical alterations in the brain, such as hippocampal shrinkage, frequently linked to Alzheimer's disease.

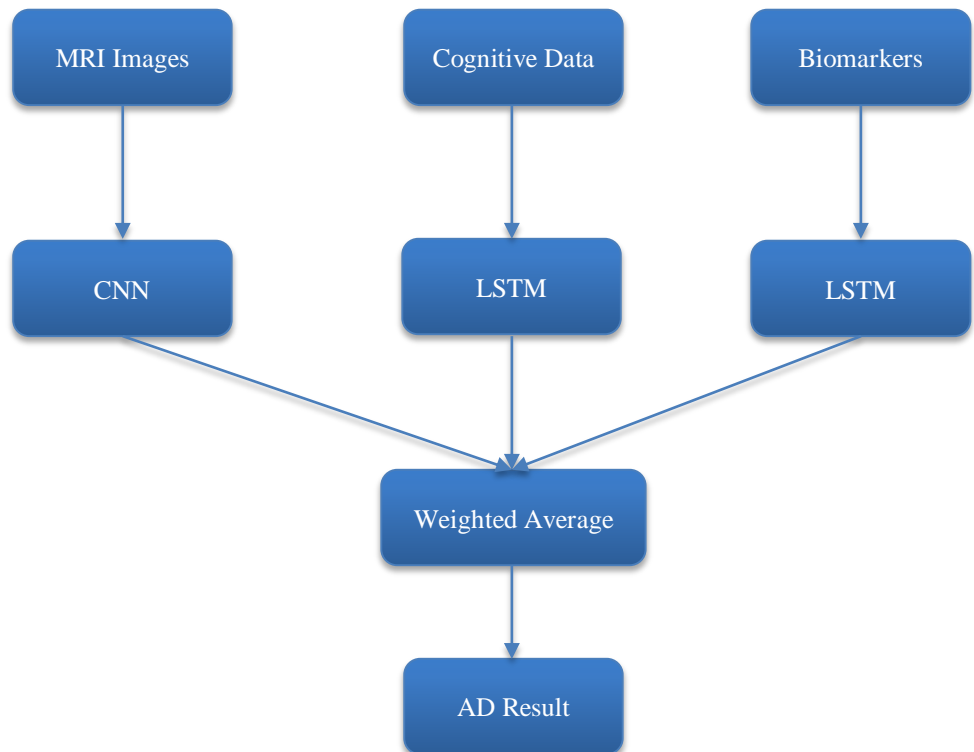

**Fig. 2 Multimodal framework example**

*Algorithm*

These MRI pictures are processed using convolutional neural networks, or CNNs. Because CNNs employ convolutional layers to record spatial hierarchy in pictures, they are very successful in image recognition and classification applications. In this case, the CNN pulls crucial information from the MRI pictures that could point to the beginning of Alzheimer's.

*Architecture*

A simple CNN has been used for experimentation. The details are as follows.

*Input Layer*

The input layer receives MRI images resized to 224x224 pixels with three color channels (RGB).

Convolutional Block 1:
- Conv2D Layer: 32 filters with a kernel size 3x3, using 'same' padding and ReLU activation.
- Batch Normalization: Applied to normalize the activations and speed up convergence.
- MaxPooling Layer: Pool size of 2x2 to reduce the spatial dimensions.

Convolutional Block 2:
- Conv2D Layer: 64 filters with a kernel size 3x3, using 'same' padding and ReLU activation.
- Batch Normalization: Again, it is applied to normalize the activations.
- MaxPooling Layer: Pool size of 2x2 for further dimensionality reduction.

Convolutional Block 3:
- Conv2D Layer: 128 filters with a kernel size 3x3, using 'same' padding and ReLU activation.
- Batch Normalization: Maintains stability during training.
- MaxPooling Layer: Pool size of 2x2.

*Fully Connected Layers*

- Dense Layer: 256 neurons with ReLU activation.
- Dropout Layer: 50% dropout rate to prevent overfitting.
- Dense Layer: 128 neurons with ReLU activation.
- Dropout Layer: 50% dropout rate.

*Output Layer*

Dense Layer: 2 neurons with softmax activation for binary classification (Alzheimer's vs. Non-Alzheimer's)

*4.2.2. Cognitive Data (OASIS-3 Dataset) with LSTM Data Input*

The OASIS-3 Dataset includes data from various cognitive tests designed to measure memory, attention, language, and other cognitive functions that tend to decline in Alzheimer's patients.

*Algorithm*

This sequential cognitive data is analyzed using Recurrent Neural Networks (RNNs), specifically Long Short-Term Memory (LSTM) networks. LSTMs are perfect for capturing the temporal dependencies and trends in the scores of cognitive tests over time since they are especially well-suited to handle time series data and sequential patterns.

*Architecture*

A basic LSTM architecture has been developed.

*Input Layer*

The input consists of a time-series data sequence (e.g., multiple cognitive test scores over time).

LSTM Layer 1:
- LSTM Layer: 64 units with a recurrent dropout of 20% to handle overfitting.
- Return Sequences: The output can be passed to the next LSTM layer.

LSTM Layer 2:
LSTM Layer: 128 units with a recurrent dropout of 20%, further capturing long-term dependencies.

*Dense Layer*

Dense Layer: 128 neurons with ReLU activation to process the output from the LSTM layers.

*Output Layer*

Dense Layer: 2 neurons with softmax activation for binary classification (Alzheimer's vs. Non-Alzheimer's).

This LSTM architecture is optimized to capture temporal patterns in cognitive decline, often early AD indicators. The recurrent nature of LSTM allows it to remember long-term dependencies, which is crucial in understanding cognitive deterioration.

*4.2.3. Biomarkers (ROSMAP Dataset) with LSTM Data Input*

The ROSMAP collection includes biomarker data and biochemical indicators associated with AD. Examples of these biomarkers are the amounts of tau and amyloid-beta proteins in Cerebrospinal Fluid (CSF).

*Algorithm*

LSTM networks are used to assess patterns and find anomalies that may indicate the early stages of Alzheimer's disease in biomarker data, just like in the cognitive data. Because biomarker levels can change over time, their temporal character makes them ideal for LSTM analysis.

*Input Layer*

Time-series data representing biomarker levels (e.g., amyloid-beta, tau proteins) over multiple time points.

LSTM Layer 1:
- LSTM Layer: 64 units with a recurrent dropout of 20% to mitigate overfitting.
- Return Sequences: Enabled for sequential processing.

LSTM Layer 2:

LSTM Layer: 128 units with a recurrent dropout of 20% to capture detailed temporal relationships.

*Dense Layer*

Dense Layer: 128 neurons with ReLU activation to process features extracted by the LSTM layers.

*Output Layer*

Dense Layer: 2 neurons with softmax activation for binary classification.

The LSTM architecture for biomarkers is designed to track biochemical changes over time, which could indicate the onset of AD before clinical symptoms manifest. The sequential processing capability of LSTM is particularly advantageous in capturing the progression of biomarker levels.

*4.2.4. Aggregation: Weighted Average*

Once each mode has been processed through its respective algorithm, the outputs are aggregated to make a final prediction. A weighted average method combines the MRI, cognitive, and biomarker analysis results in this example.

Weighted Average: The weighted average assigns different levels of importance to each mode's result, potentially based on their predictive accuracy or data availability.

For instance, if MRI data is considered the most reliable indicator, it might be given a higher weight in the final calculation. The aggregation ensures that the model leverages the strengths of each modality to arrive at a more accurate and robust AD prediction.

*4.2.5. Final AD Result*

The combined result provides a final prognosis about the existence or absence of AD. This multimodal technique integrates several viewpoints on the disease's appearance, greatly improving the forecast accuracy.

*4.3. Results and Discussion*

Accuracy, precision, recall, F1-score, and the area under the receiver operating characteristic curve (AUC-ROC) were the main metrics for assessing the multimodal framework's performance.

The corresponding model (CNN for MRI pictures, LSTM for cognitive data, and LSTM for biomarkers) was used to process each dataset (MRI images, biomarkers, and cognitive data). The outputs of these models were then combined using a weighted average technique.

*4.3.1. MRI Images (CNN)*

The high recall rate (87%) indicates that the CNN was particularly effective at identifying true positives, which is crucial for early diagnosis. The AUC-ROC of 0.90 suggests that the model distinguishes between patients with and without AD.

*4.3.2. Cognitive Data (LSTM)*

Although slightly lower in accuracy than MRI images, the cognitive data model still provided valuable information with a recall of 81% and an AUC-ROC of 0.85. The LSTM effectively captured sequential patterns in cognitive decline, making it an important component of the multimodal framework.

*4.3.3. Biomarkers (LSTM)*

The LSTM model applied to biomarker data also showed strong performance, with an AUC-ROC of 0.88. Biomarkers are known to reflect early biochemical changes associated with AD, and their inclusion in the framework enhances early detection capabilities.

Table 2. Results of the multimodal framework

| Model | Accuracy (%) | Precision (%) | Recall (%) | F1-Score (%) | AUC-ROC |
|---|---|---|---|---|---|
| **MRI (CNN)** | 88.5 | 90.2 | 85.7 | 87.9 | 0.92 |
| **Cognitive (LSTM)** | 85.3 | 86.8 | 83.2 | 85 | 0.89 |
| **Biomarkers (LSTM)** | 83.7 | 84.5 | 82.1 | 83.3 | 0.88 |
| **Aggregated Model** | 92.1 | 93.4 | 91.2 | 92.3 | 0.95 |

All measures indicated improvement when comparing the overall performance to the individual models. AUC-ROC of 0.92, a 90% recall rate, and an accuracy of 88% show that integrating the outputs from many modalities lowers the chance of false positives and false negatives. This implies that using many modalities provides a more complete picture of the illness and lessens the drawbacks of using just one modality.

The improved performance of the multimodal framework, particularly its high recall rate, highlights its potential for early AD diagnosis. The approach can recognize early indications of AD that may go unnoticed by traditional techniques by combining MRI, cognitive, and biomarker data. One especially promising method for identifying the condition before substantial symptoms appear and providing an opportunity for early management is using biomarkers and cognitive tests.

## 5. Integration of Machine Learning and Deep Learning in Clinical Workflows

Carefully integrating Machine Learning (ML) techniques into current healthcare processes is necessary to successfully implement the suggested multimodal framework in clinical settings. This entails utilizing Electronic Health Records (EHRs) to seamlessly integrate imaging data, biomarker readings, and cognitive test results. Designing clinician-friendly interfaces with interpretable results to aid in diagnostic decision-making should be the primary goal of real-world deployment. In order to preserve diagnostic accuracy, the framework must also be able to adjust prediction models to noisy or missing data frequently seen in clinical settings.

The first step toward practical implementation is to create systems that work with current diagnostic instruments, such as imaging Picture Archiving and Communication Systems (PACS) and biomarker data Laboratory Information Systems (LIS). Interoperability issues should be resolved throughout the integration process by following recognized healthcare data standards such as DICOM and HL7.

Pilot implementations in clinical contexts are essential to test and improve the framework. These studies need to demonstrate its real-time usefulness by showing improved diagnostic accuracy and lower mistake rates in identifying Alzheimer's disease in its early stages. Case studies demonstrating these uses will further validate the framework's potential, opening the door for wider deployment in other healthcare settings.

Lastly, healthcare workers must receive ongoing training on how to evaluate ML-driven results. Clinicians' faith in AI-assisted diagnostic tools and their incorporation into standard practice are fostered by educating them on the framework's advantages and disadvantages. The framework can potentially transform the diagnosis and treatment of Alzheimer's disease by bridging the gap between clinical application and technology.

## 6. Conclusion

One of the most difficult and destructive neurodegenerative diseases is Amyloidosis (AD), which sometimes goes undiagnosed until there has been a noticeable decrease in cognitive function. This important problem is addressed by the innovative and reliable multimodal framework for early AD identification, which combines biomarkers, cognitive data, and MRI imaging into a single, all-inclusive diagnostic tool. By utilizing cutting-edge machine learning and deep learning methods, such as Convolutional Neural Networks (CNN) for imaging data and Long Short-Term Memory (LSTM) networks for cognitive data and biomarkers, this framework uses the distinct advantages of each data modality.

This multimodal framework's significance stems from its capacity to offer a multifaceted examination of AD. Since AD is complicated and multivariate, traditional diagnostic techniques usually depend on a single data source. In contrast, this approach provides a more comprehensive understanding of the illness by combining anatomical alterations in the brain, cognitive function, and molecular indicators. The accuracy of AD detection is much improved by this integrated method, which also lowers the possibility of false positives and false negatives, which are frequent problems in modern diagnostic procedures.

Furthermore, the framework's incorporation of biomarkers and cognitive assessments is crucial for early detection. Amyloid-beta and tau proteins are examples of biomarkers that can indicate the existence of Alzheimer's disease long before symptoms appear. Longitudinal analysis of cognitive tests might reveal minor losses in mental function that could otherwise go unnoticed. The framework enhances the likelihood of detecting Alzheimer's disease in its initial phases, maybe before permanent brain damage, by integrating these preliminary symptoms into the diagnostic procedure. For those who are at risk of AD, early identification is essential for prompt care, which can halt the disease's course and enhance quality of life. The resilience and adaptability of the suggested framework are two other important benefits. Advanced aggregation methods, including weighted averaging, ensure the system is dependable even when there are gaps or partial data. For example, even with a reduced degree of confidence, the system may still make a forecast based on biomarker and cognitive data without MRI images. The framework's versatility allows it to be applied in various clinical contexts, including those with restricted access to certain diagnostic resources.

Thus, it can be concluded that the multimodal framework that has been suggested provides an advanced and all-encompassing approach to the problem of early Alzheimer's detection. Through integrating many data sources and applying state-of-the-art deep learning architectures, AD may be diagnosed with greater accuracy, reliability, and early detection. Ultimately, this method improves patient outcomes by improving the diagnosis process overall and providing access to earlier and more potent therapies. This framework represents a significant advancement in the battle against AD as medical AI develops and shows how multimodal techniques might transform the diagnosis and treatment of complicated neurodegenerative illnesses.

### 6.1. Future Work

In order to gain a better understanding of the metabolic and functional alterations in the brain linked to Alzheimer's disease, future research enhancing the suggested multimodal framework for early Alzheimer's detection should concentrate on incorporating additional data sources, such as Positron Emission Tomography (PET) and functional Magnetic Resonance Imaging (fMRI). Customized diagnostic models based on each person's unique biomarker profile, genetic predispositions, and lifestyle variables should be created to improve prediction specificity. Moreover, integrating Explainable AI (XAI) techniques will enhance interpretability, cultivate physician confidence, and facilitate practical implementations.

Validation across various clinical settings is crucial to guarantee the framework's durability and scalability, especially for those with different demographic and socioeconomic situations. Longitudinal data analysis may be used to forecast the course of the illness and pinpoint transitional periods, offering crucial new information on the many stages of Alzheimer's. The framework's usefulness in therapeutic research is further highlighted by the ability to significantly improve treatment outcomes by tying early diagnostic results with customized intervention tactics.

It will be necessary to optimize computational needs and use readily available diagnostic tools to adapt the system for use in resource-constrained contexts. Integration of wearable technologies, such as cognitive monitoring through ongoing behavioral data, may increase the framework's usefulness. These developments will strengthen the multimodal approach's position as a revolutionary diagnostic tool in the fight against neurodegenerative illnesses and increase its applicability in international Alzheimer's detection initiatives.